\documentclass[letterpaper]{article} 
\usepackage{aaai2026}  
\usepackage{times}  
\usepackage{helvet}  
\usepackage{courier}  
\usepackage[hyphens]{url}  
\usepackage{graphicx} 
\usepackage{natbib}  
\usepackage{caption} 
\usepackage{algorithm}
\usepackage{algorithmic}
\usepackage{amsmath}
\usepackage{amssymb}
\usepackage{booktabs}
\usepackage{pifont}
\usepackage{tabularx}
\usepackage{colortbl}
\usepackage{xcolor}
\definecolor{mediumblue}{RGB}{0, 87, 183}       
\newcommand{\colorurl}[1]{\textcolor{mediumblue}{\url{#1}}}

\usepackage{newfloat}
\usepackage{listings}
\DeclareCaptionStyle{ruled}{labelfont=normalfont,labelsep=colon,strut=off} 
\floatstyle{ruled}
\newfloat{listing}{tb}{lst}{}
\floatname{listing}{Listing}
\title{Do We Need Perfect Data? \\ Leveraging Noise for Domain Generalized Segmentation}
\author{
    Taeyeong Kim, SeungJoon Lee, Jung Uk Kim$^*$, MyeongAh Cho\thanks{Corresponding author.}
}
\affiliations{
     Kyung Hee University, Republic of Korea\\
    \{rlaxo0511, diplomat3334, ju.kim, maycho\}@khu.ac.kr
}

\usepackage{bibentry}

\begin{document}
\maketitle

\begin{abstract}
Domain generalization in semantic segmentation faces challenges from domain shifts, particularly under adverse conditions. While diffusion-based data generation methods show promise, they introduce inherent misalignment between generated images and semantic masks. This paper presents \textbf{FLEX-Seg} (FLexible Edge eXploitation for Segmentation), a framework that transforms this limitation into an opportunity for robust learning. FLEX-Seg comprises three key components: (1) \textbf{Granular Adaptive Prototypes} that captures boundary characteristics across multiple scales, (2) \textbf{Uncertainty Boundary Emphasis} that dynamically adjusts learning emphasis based on prediction entropy, and (3) \textbf{Hardness-Aware Sampling} that progressively focuses on challenging examples. By leveraging inherent misalignment rather than enforcing strict alignment, FLEX-Seg learns robust representations while capturing rich stylistic variations. Experiments across five real-world datasets demonstrate consistent improvements over state-of-the-art methods, achieving 2.44\% and 2.63\% mIoU gains on ACDC and Dark Zurich. Our findings validate that adaptive strategies for handling imperfect synthetic data lead to superior domain generalization. Code is available at \colorurl{https://github.com/VisualScienceLab-KHU/FLEX-Seg}.

\end{abstract}


\section{Introduction}

Semantic segmentation, which assigns semantic labels to each pixel in an image, is fundamental to many computer vision applications, especially autonomous driving~\cite{Shelhamer2014FullyCN, chen2017deeplab, badrinarayanan2017segnet}. However, models trained on one domain often fail on unseen domains due to \emph{domain shift} caused by variations in weather, lighting, and imaging conditions~\cite{li2023domain, hoffman2018cycada}, creating a \emph{domain gap}~\cite{luo2019taking} that severely limits real-world deployment. While Domain Adaptive Semantic Segmentation addresses this by using unlabeled target domain data during training~\cite{ganin2015unsupervised, hoffman2018cycada}, collecting such data is often impractical~\cite{li2017deeper}. This motivates Domain Generalized Semantic Segmentation (DGSS), which trains models to generalize to unseen domains using only source domain data~\cite{zhou2022domain, carlucci2019domain}, eliminating any need for target domain data access (neither images nor labels). Thus, DGSS is more suitable for real-world applications where deployment environments are unknown or constantly changing.

Recent studies in DGSS have explored diverse approaches for learning domain-invariant features. Early methods focused on normalization techniques~\cite{pan2018two, choi2021robustnet} and style randomization~\cite{yue2019domain, kim2023texture} to reduce domain-specific biases. Subsequently, data augmentation via image-to-image translation has been widely adopted to simulate diverse domain conditions~\cite{li2023intra, huang2021fsdr}. Recently, diffusion-based generative methods have set new benchmarks. Particularly, DGInStyle~\cite{jia2024dginstyle} leverages pre-trained latent diffusion models to synthesize diverse yet semantically consistent images, significantly improving generalization by exploiting large-scale generative priors, although they often introduce spatial misalignments at fine-grained levels such as object boundaries.

\begin{figure}[t]
    \centering
    \includegraphics[width=0.47\textwidth]{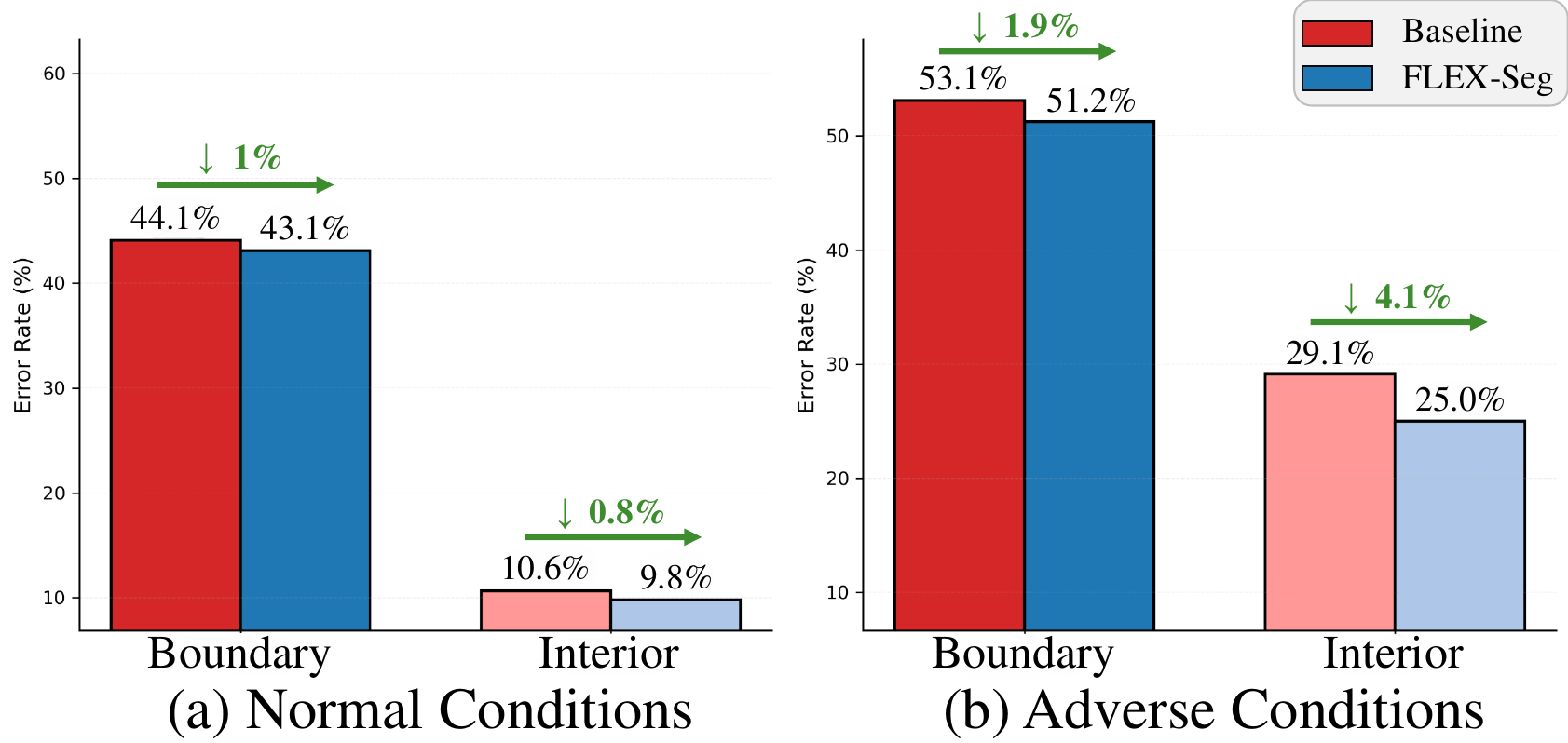} 
    \caption{Error rate analysis of boundary vs. interior regions when training with diffusion-generated synthetic data. Under both (a) normal and (b) adverse conditions, boundary regions consistently exhibit higher error rates than interior regions, with the disparity becoming more pronounced in adverse conditions. \textbf{FLEX-Seg} effectively reduces errors in both regions across all scenarios, demonstrating that our boundary-focused approach is crucial for robust domain generalization.}
    \label{fig:intro}
\end{figure}

\begin{figure}[t]
    \centering
    \includegraphics[width=0.47\textwidth]{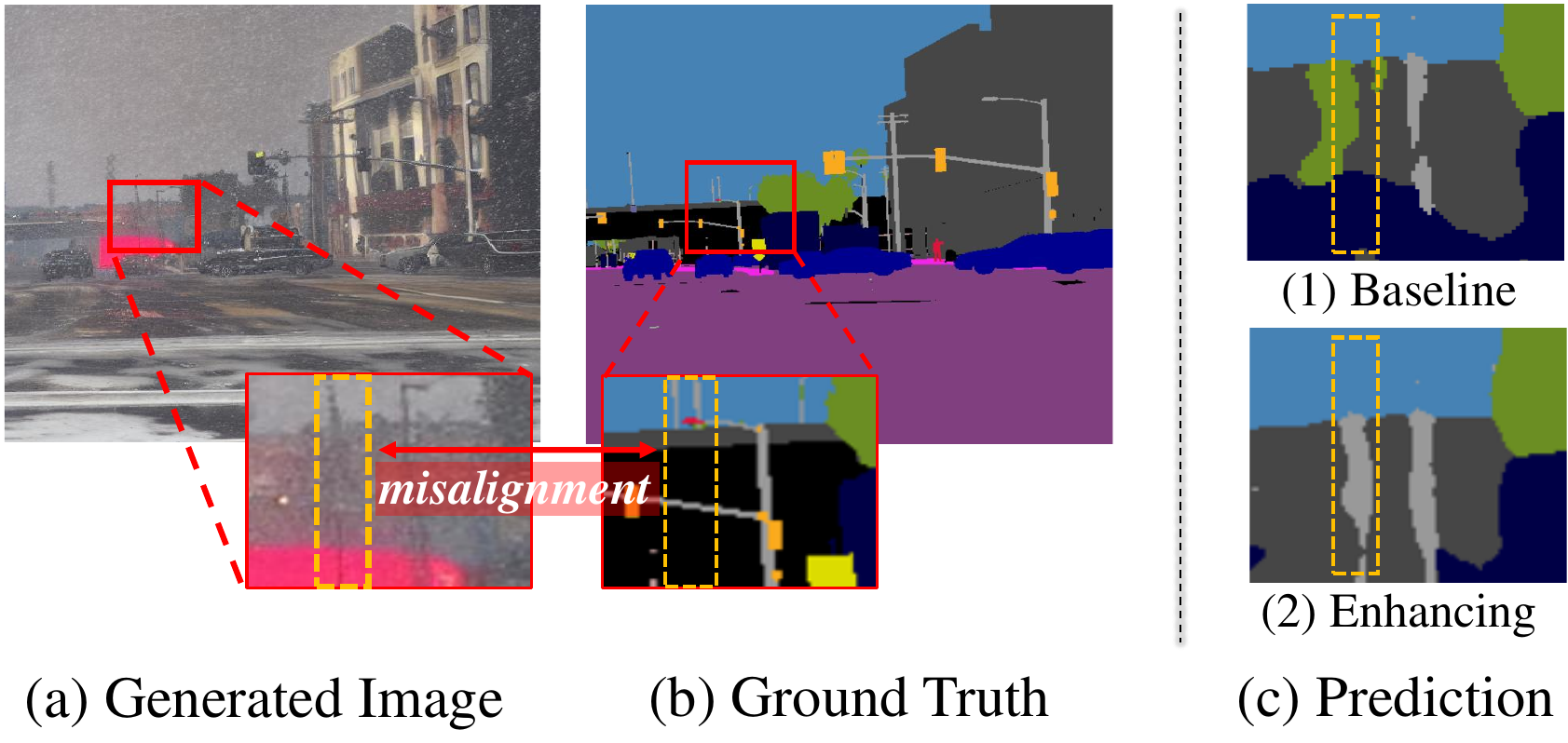} 
    \caption{Column 1: Generated image and corresponding ground truth mask. Column 2: Model predictions on (a) generated image using different approaches. \textbf{Better view in zoom.}}
    \label{fig:motiv}
\end{figure}

However, a fundamental challenge remains: accurately predicting object boundaries is crucial for semantic segmentation, as boundaries define the shape and structure of objects~\cite{marmanis2018classification, li2020improving}. This challenge is particularly critical in DGSS, where models must generalize boundary representations to diverse boundary appearances across unseen domains without access to target domain data, making them more vulnerable to cross-domain variations in boundary visibility and structure. The inherent characteristics of adverse conditions—such as fog, snow, and nighttime scenarios—make this challenge even more severe, as these conditions naturally produce blurred boundaries and reduced visibility that obscure the clear delineation between objects~\cite{lei2020semantic, narayanan2021shape}. As shown in Fig.~\ref{fig:intro}, boundary regions exhibit significantly higher error rates than interior regions, with errors becoming even larger under adverse conditions.
This issue is exacerbated when utilizing generated synthetic datasets for DGSS, where the generation process often fails to achieve \emph{perfect pixel-wise alignment}. In contrast to real datasets—where semantic mask labels are typically derived from real images—synthetic data pipelines generate images from semantic masks. This reversed generation process inherently introduces spatial misalignments between the synthesized images and their associated labels. As illustrated in the first column of Fig.~\ref{fig:motiv}, fine-grained structures in the generated images often fail to align precisely with their semantic masks. Such inconsistencies can mislead the model into focusing on resolving local ambiguities rather than learning meaningful cross-domain variations.

Existing boundary-aware methods~\cite{zhang2024boundary, borse2021inverseform, ngoc2021introducing, liu2021bapa} are fundamentally limited in handling the inherent misalignment present in synthetic data. These approaches are predicated on the assumption of perfect spatial correspondence between image structures and semantic mask annotations—an assumption that does not hold in synthetically generated datasets. Therefore, addressing the dual challenges of boundary precision from synthetic data misalignment requires a new approach that learns robust representations from imperfect alignments while capturing rich stylistic variations in less constrained regions.

To address these challenges, we propose \textbf{FLEX-Seg} (\textbf{FL}exible \textbf{E}dge e\textbf{X}ploitation for \textbf{Seg}mentation), a novel framework that transforms the inherent misalignment in synthetic data into an opportunity for learning more robust and domain-invariant representations. FLEX-Seg comprises three carefully designed components that synergistically balance precise boundary delineation with rich stylistic learning. (1) \textbf{Granular Adaptive Prototypes (GAP)} captures boundary characteristics at different thickness levels by organizing prototypes according to both semantic class and boundary granularity. By constructing a structured prototype bank across three granularity levels—from thin boundaries in distant objects to thick regions in nearby ones—GAP learns domain-invariant boundary representations while maintaining the natural scale variations of object boundaries. (2) \textbf{Uncertainty Boundary Emphasis (UBE)} modulates supervision strength based on prediction entropy. This mechanism dynamically amplifies the contribution of uncertain pixels—typically at misaligned boundaries and ambiguous regions—while maintaining standard gradients for confident predictions. By directing learning capacity toward these challenging areas, UBE enhances boundary discrimination without overfitting to synthetic data artifacts, improving robustness across diverse domains. (3) \textbf{Hardness-Aware Sampling (HAS)} optimizes training efficiency by progressively focusing on challenging examples. Through sigmoid decay scheduling that gradually transitions from random to loss-based sampling, HAS ensures efficient resource allocation to the most informative samples, particularly those with complex structures or adverse conditions.

Comprehensive domain generalization experiments across five real-world datasets demonstrate that FLEX-Seg consistently outperforms existing state-of-the-art methods. Our approach achieves significant improvements on challenging domains with adverse conditions, gaining 2.44\% and 2.63\% mIoU on ACDC (fog, rain, snow, and nighttime scenarios) and Dark Zurich (nighttime driving conditions) respectively. These results validate our hypothesis that leveraging inherent misalignment in synthetic data rather than enforcing strict alignment leads to learn more robust and transferable representations, ultimately leading to improved generalization across unseen domains.

\begin{figure*}[t]
    \centering  
    \includegraphics[width=1.0\textwidth]{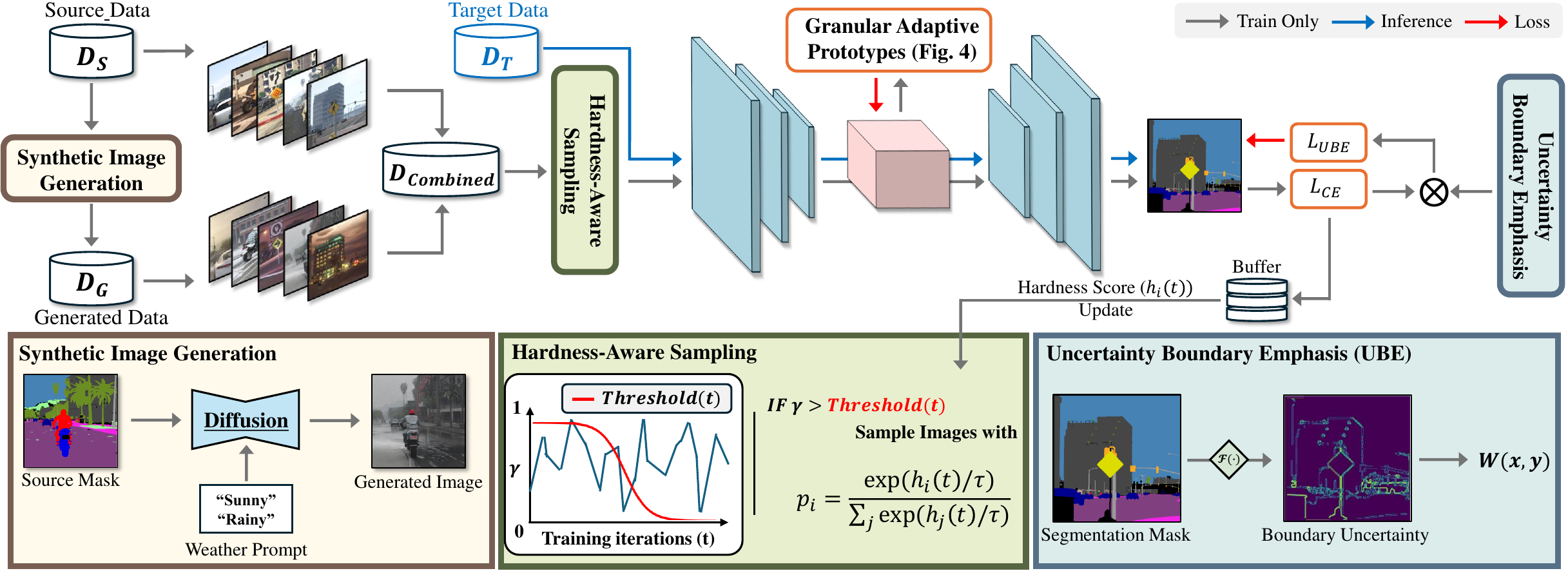}
    \caption{Overview of our FLEX-Seg framework. The framework integrates three key components: Granular Adaptive Prototypes for learning domain-invariant boundary representations, Uncertainty Boundary Emphasis for dynamically emphasizing challenging regions, and Hardness-Aware Sampling for efficient training on difficult examples. These components work synergistically to improve boundary precision across diverse domains.}
    \label{fig:overview}
\end{figure*}

\section{Related Work}

\subsection{Dataset Generation for Domain Generalization}
Recent approaches improve training data quality using diffusion models for diverse synthetic generation. CLOUDS~\cite{benigmim2024collaborating} combines foundation models for enhanced diversity in domain generalization. ALDM~\cite{li2024adversarial} introduces adversarial supervision to preserve layout fidelity in layout-to-image diffusion models. DGInStyle~\cite{jia2024dginstyle} uses latent diffusion with style swapping to create semantically consistent images across domains. Despite these efforts to improve generation quality, these methods often cause boundary misalignment between generated images and semantic masks, an issue that remains underexplored; our approach leverages this misalignment for robust representations.

\subsection{Domain Generalized Semantic Segmentation}
Domain generalization in semantic segmentation aims to train models that perform well on unseen domains without access to target data. Early methods used normalization techniques like instance normalization~\cite{pan2018two} and whitening transformation~\cite{li2017universal, choi2021robustnet} to reduce domain-specific style variations. Domain randomization simulates diverse conditions via style transfer and augmentation~\cite{yue2019domain, kim2023texture}. Recent advances leverage modern architectures like DAFormer~\cite{hoyer2022daformer} for Transformer-based training, HRDA~\cite{hoyer2023domain} for multi-resolution fusion. Further extensions include language-guided approaches~\cite{fahes2024simple} and semantic consistency learning~\cite{niu2025exploring}.

\section{Motivation}

We first investigate the effect of different boundary handling strategies to address the misalignment issue between the synthesized images and their associated labels in synthetic training data. Since misalignment primarily manifests at object boundaries where generated textures fail to precisely follow semantic masks, we hypothesize that focusing on these critical regions could improve model robustness. The most intuitive approach is to apply stronger weights to boundary regions during training, forcing the model to pay more attention to these challenging areas despite their imperfect alignment. We identify boundary regions $B$ through morphological operations on semantic masks:

\begin{equation}
B = \text{Dilate}(M, k_d) - \text{Erode}(M, k_e),
\label{eq:boundary}
\end{equation}
where $M$ is the semantic mask, and $k_d$, $k_e$ are dilation and erosion kernel sizes respectively.

We then incorporate enhanced weighting into the training loss to explicitly emphasize learning on boundary regions:
\begin{equation}
\mathcal{L}_{\text{boundary}} = \sum_{(x,y)} W(x,y) \cdot \mathcal{L}_{\text{CE}}(x,y),
\end{equation}
where $W(x,y) = \alpha > 1$ if $(x,y) \in B$, and $1$ otherwise. The parameter $\alpha$ adjusts the loss weight for boundary pixels, guiding the model to focus more on accurately capturing object boundaries.

As shown in the second column of Fig.~\ref{fig:motiv}, this simple boundary-enhancing approach effectively captures object characteristics even in misaligned synthetic data. The model trained with boundary emphasis ($\alpha = 5$) shows more robust predictions compared to the baseline, successfully delineating object structures despite the substantial misalignment between generated images and semantic masks.

This observation inspired us to design more sophisticated modules that not only emphasize boundaries but also adapt to their inherent uncertainty and multi-scale characteristics. Rather than using a fixed weight $\alpha$, our GAP module learns boundary representations across multiple granularities, while UBE dynamically adjusts weights based on prediction uncertainty.
Additional motivation experiments exploring various boundary weighting strategies (ignore, reduce, threshold) can be found in the supplementary material.

\section{Method}
\subsection{DG for Semantic Segmentation}
The goal of domain generalization in semantic segmentation is to learn a model using only labeled source domain data that can perform well on unseen target domains. Formally, let the source domain be denoted by $\mathcal{D}_S = \left\{(x_i^S, y_i^S)\right\}_{i=1}^{N_S}$, where $x_i^S$ represents an image in the source domain and $y_i^S$ is its pixel-level semantic label. Our objective is to learn a segmentation model that generalizes to unseen target domains $\mathcal{D}_T = \left\{ x_i^T \right\}_{i=1}^{N_T}$, where $x_i^T$ denotes an image in the target domain and labels are unavailable during training, reflecting the typical domain generalization setup.

\subsection{FLEX-Seg Framework}

The FLEX-Seg framework addresses boundary precision challenges in diffusion-based domain generalization through two sophisticated modules: Granular Adaptive Prototypes (GAP) and Uncertainty Boundary Emphasis (UBE), alongside Hardness-Aware Sampling (HAS). Fig.~\ref{fig:overview} illustrates the comprehensive architecture of our approach.

Initially, diverse synthetic images are generated using diffusion models on source dataset $\mathcal{D}_S$ and its semantic masks, forming augmented dataset $\mathcal{D}_G$ and combining it with $\mathcal{D}_S$ to create unified training corpus $\mathcal{D}_{combined} = \mathcal{D}_S \cup \mathcal{D}_G$.

The training process incorporates a dual-pathway boundary refinement strategy. GAP constructs a two-dimensional prototype bank that captures boundary characteristics across class semantics and granular intensity levels, enabling domain-invariant representation learning through contrastive alignment, while UBE dynamically adjusts pixel-wise loss contributions based on prediction uncertainty, providing adaptive emphasis on challenging boundary regions without manual hyperparameter tuning.

Integration of prototype-based learning and uncertainty-driven weighting provides a robust foundation for domain generalization, effectively handling diverse visual conditions and complex boundary structures across domains.

\subsection{Granular Adaptive Prototypes (GAP)}

\begin{figure}[t]
    \centering
    \includegraphics[width=0.47\textwidth]{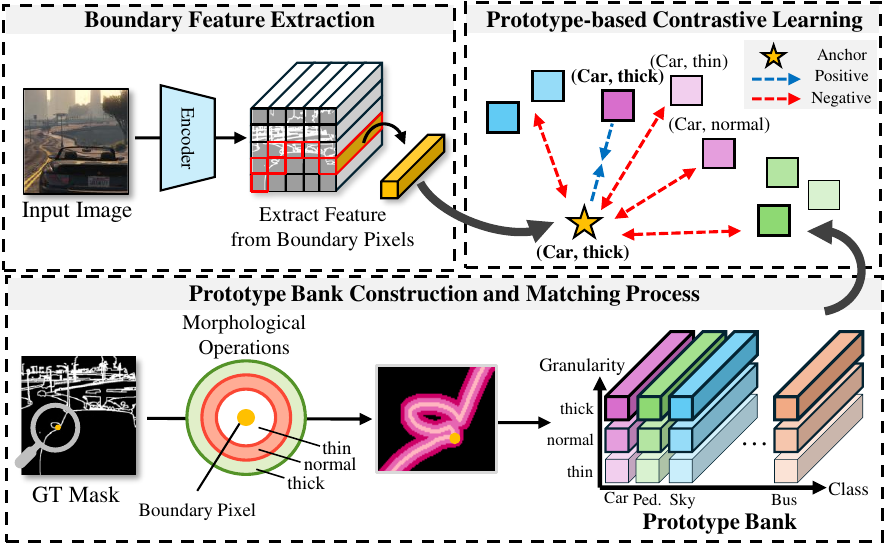} 
    \caption{Illustration of GAP. Boundary features are extracted at multiple granularities, assigned to corresponding class-granularity prototypes, and refined through contrastive learning to achieve domain-invariant representations.}
    \label{fig:mbpc}
\end{figure}

The GAP module addresses domain-invariant boundary representation learning through a novel two-dimensional coordinate system, as illustrated in Fig.~\ref{fig:mbpc}. This approach stems from the inherent scale variability of semantic boundaries, where distant small objects exhibit thin boundaries while nearby large objects present thick boundary regions. Additionally, boundary pixels exhibit two distinct types of variations: geometric variations (how thick or thin the boundary appears) and stylistic variations (how the boundary appears under different environmental conditions).

\noindent\textbf{Class-Shape Token Coordinate System.} We formalize this concept by decomposing boundary characteristics into orthogonal dimensions. Each boundary pixel $p_i$ is represented as a point in a coordinate space $(c_i, g_i)$, where the class token $c_i$ captures the semantic class identity, and the shape token $g_i$ encodes geometric attributes such as boundary thickness.

To extract multi-granular boundary representations, we downsample the ground truth mask $M$ from resolution $(H, W)$ to match feature maps at $(H_f, W_f)$ using stride $s = H / H_f$ to obtain $M_d$. We then generate three boundary masks through morphological operations:

\begin{equation}
B_g = \text{Dilate}(M_d, k_g) \ominus \text{Erode}(M_d, k_g)
\end{equation}
where $k_g$ represents kernel sizes for generating thin, medium, and thick boundary granularities respectively.

This decomposition enables systematic analysis of boundary characteristics by separating semantic and geometric properties that should remain consistent across domains from stylistic variations due to environmental conditions.

\noindent\textbf{Prototype Bank Construction.} We construct a prototype bank $\mathcal{P} = \{p_{c,g}\}$ where $p_{c,g} \in \mathbb{R}^{256}$ represents the prototype for class $c$ with granularity $g$, resulting in $C \times 3$ prototypes ($C$ classes × 3 granularities). The prototype bank serves as a structured memory that captures representative boundary features for each class-granularity combination, enabling consistent boundary representations across different domains.

Prototypes are updated via pixel-wise momentum updates during training:
\begin{equation}
p_{c,g} \leftarrow m \cdot p_{c,g} + (1-m) \cdot f_{c,g}
\end{equation}
where $m \in [0,1]$ is the momentum factor, and $f_{c,g}$ is the feature vector of a boundary pixel belonging to class $c$ with granularity $g$. This pixel-wise update strategy allows for fine-grained prototype refinement while maintaining stability through the momentum mechanism.

\noindent\textbf{Prototype-based Contrastive Learning.} The InfoNCE loss enforces that boundary pixels with identical class-granularity characteristics converge regardless of domain origin. To address class imbalance, we incorporate adaptive weighting based on prototype exposure frequency:

{\footnotesize
\begin{equation}
\mathcal{L}_{GAP} = -\frac{1}{N} \sum_{i=1}^{N} w_{c_i,g_i} \cdot \log \frac{e^{\langle f_i, p_{c_i,g_i} \rangle / \tau}}{\sum_{(c',g') \in \mathcal{P}} e^{\langle f_i, p_{c',g'} \rangle / \tau}}
\end{equation}
}where $\tau$ is the temperature parameter, $N$ is the number of boundary pixels in the batch, $\mathcal{P}$ represents all class-granularity combinations in the prototype bank, and $\langle \cdot, \cdot \rangle$ denotes the cosine similarity. The weight $w_{c_i,g_i}$ balances the contribution of each class-granularity combination.

For each prototype $p_{c,g}$, we maintain its update frequency $u_{c,g}$. The imbalance-aware weight is computed as:

\begin{equation}
w_{c,g} = \frac{\max(u) + 1}{u_{c,g} + 1} \cdot \frac{1}{Z}
\end{equation}
where $\max(u)$ denotes the highest update frequency among all prototypes, and $Z$ is a normalization factor ensuring $\max(w_{c,g}) = 1$. This formulation assigns higher weights to under-represented class-granularity combinations, effectively balancing the learning across all boundary types.

This comprehensive approach enables domain-invariant boundary learning while addressing the natural imbalance in boundary occurrence frequencies.

\subsection{Uncertainty Boundary Emphasis (UBE)}

The UBE module dynamically emphasizes challenging boundary regions by leveraging prediction uncertainty, eliminating the need for manual hyperparameter tuning across diverse domains. Unlike fixed weighting schemes, this approach automatically adapts to varying difficulty levels where nighttime or adverse weather conditions present more uncertain boundary predictions.

\noindent\textbf{Entropy-based Dynamic Weighting.} We compute prediction entropy as an uncertainty indicator, where higher values correspond to ambiguous boundary regions. For each pixel location $(x,y)$, the entropy is calculated as $H_{x,y} = -\sum_{c=1}^{C} p_c(x,y) \log p_c(x,y)$, where $p_c(x,y)$ is the predicted probability for class $c$ at location $(x,y)$. The adaptive weighting is applied exclusively to boundary regions while maintaining unit weight for interior pixels:

{\footnotesize
\begin{equation}
w(x,y) = \begin{cases}
1 + \alpha \cdot \text{sigmoid}\left(\frac{H_{x,y} - \mu_H}{\sigma_H + \epsilon}\right), & \text{if } (x,y) \in B \\
1, & \text{otherwise}
\end{cases}
\end{equation}
}where $B$ represents the boundary mask, $\mu_H$ and $\sigma_H$ are batch-wise entropy statistics within boundary regions, and $\alpha$ controls maximum weight amplification.

The uncertainty-adaptive weights are incorporated into the cross-entropy loss:

\begin{equation}
\mathcal{L}_{UBE} = \frac{1}{N} \sum_{(x,y)} w(x,y) \cdot \mathcal{L}_{CE}(x,y)
\end{equation}
where $N$ is the total number of pixels and $\mathcal{L}_{CE}(x,y)$ is the cross-entropy loss at pixel location $(x,y)$. This mechanism automatically emphasizes difficult boundary regions while preserving stable learning for confident predictions, requiring only a single stable hyperparameter $\alpha$ across different environmental conditions. 

\noindent\textbf{Loss Function Integration.} The complete training objective combines the UBE loss and GAP loss for comprehensive boundary refinement:

\begin{equation}
\mathcal{L}_{total} = \mathcal{L}_{UBE} + \lambda_{gap} \cdot \mathcal{L}_{GAP}
\end{equation}

While GAP enables domain-invariant boundary learning through prototype-based contrastive learning, UBE provides adaptive emphasis on uncertain regions. Together, these mechanisms create a synergistic effect: GAP ensures consistent boundary representations across domains by learning from multi-granular prototypes, while UBE dynamically adjusts the learning focus based on prediction confidence. This integrated approach allows the model to effectively handle both the inherent misalignment in synthetic data and the varying difficulty levels across different environmental conditions, resulting in more robust domain generalization performance.

\subsection{Hardness-Aware Sampling (HAS)}

To optimize training efficiency, we employ HAS that dynamically prioritizes difficult examples during training. Our approach balances exploration and exploitation by gradually transitioning from random to loss-based sampling as training progresses.

For each training image $i$, we maintain a hardness score $h_i(t)$ reflecting its difficulty level at updating step t. This score is updated periodically using exponential moving average (EMA):

\begin{equation}
h_i(t) = \beta \cdot h_i(t-1) + (1 - \beta) \cdot L\bigl(f_\theta(x_i), y_i\bigr)
\end{equation}
where $L$ represents the accumulated loss value, $f_\theta$ is the segmentation model with parameters $\theta$, and $\beta \in (0,1)$ is the EMA decay factor. Periodic updates provide more stable hardness estimates compared to per-iteration updates.

\begin{table*}[t]
    \centering
    \footnotesize 
    \begin{tabularx}{\textwidth}{l||>{\centering\arraybackslash}X>{\centering\arraybackslash}X>{\centering\arraybackslash}c|>{\centering\arraybackslash}X>{\centering\arraybackslash}X>{\centering\arraybackslash}X>{\centering\arraybackslash}c}
        \toprule
        \textbf{Method} & \textbf{ACDC} & \textbf{DZ} & \textbf{Avg2} & \textbf{CS} & \textbf{BDD} & \textbf{MV} & \textbf{Avg5} \\
        \midrule
        \multicolumn{1}{c||}{} & \multicolumn{7}{c}{\textbf{ResNet-101}} \\
        \midrule
        DRPC~\cite{yue2019domain} & -- & -- & -- & 42.53 & 38.72 & 38.05 & -- \\
        FSDR~\cite{huang2021fsdr} & 24.77 & 9.66 & 17.22 & 44.80 & 41.20 & 43.40 & 32.77 \\
        GTR~\cite{peng2021global} & -- & -- & -- & 43.70 & 39.60 & 39.10 & -- \\
        SAN-SAW~\cite{peng2022semantic} & -- & -- & -- & 45.33 & 41.18 & 40.77 & -- \\
        AdvStyle~\cite{zhong2022adversarial} & -- & -- & -- & 44.51 & 39.27 & 43.48 & -- \\
        SHADE~\cite{zhao2022style} & 29.06 & 8.01 & 18.54 & 46.66 & 43.66 & 45.50 & 34.58 \\
        FAMix$^*$~\cite{fahes2024simple} & 32.74 & -- & -- & 48.15 & \textbf{45.61} & \underline{52.11} & -- \\
        SCSD$^*$~\cite{niu2025exploring} & \underline{35.66} & -- & -- & \textbf{51.72} & \underline{44.67} & \textbf{56.98} & -- \\
        \midrule
        HRDA (Hoyer et al.~\citeyear{hoyer2023domain}) & 26.08 &  7.80 & 16.94 & 39.63 & 38.69 & 42.21 & 30.88 \\
        ~~+ DGInStyle~\cite{jia2024dginstyle} & 34.19 & 16.16 & 25.18 & 46.89 & 42.81 & 50.19 & 38.05 \\
        ~~++ FLEX-Seg (Ours) & \textbf{36.27} & \textbf{19.20} & \textbf{27.74} & \underline{48.32} & 44.03 & 51.13 & \textbf{39.79} \\
        \midrule
        \multicolumn{1}{c||}{} & \multicolumn{7}{c}{\textbf{MiT-B5}} \\
        \midrule
        DAFormer (Hoyer et al.~\citeyear{hoyer2022daformer}) &  38.25 & 17.45 & 27.85 &  52.65 & 47.89 &  54.66 & 42.18 \\
        ~~+ DGInStyle~\cite{jia2024dginstyle} & 44.04 & 25.58 & 34.81 & 55.31 & 50.82 & 56.62 & 46.47 \\
        ~~++ FLEX-Seg (Ours) & \textbf{46.56} & \textbf{29.51} & \textbf{38.04} & \textbf{56.84} & \textbf{52.06} & \textbf{57.93} & \textbf{48.58} \\
        \midrule
        HRDA (Hoyer et al.~\citeyear{hoyer2023domain}) & 44.04 & 20.97 & 32.51 & 57.41 &  49.11 & 61.16 &  46.54 \\
        ~~+ DGInStyle~\cite{jia2024dginstyle} & 46.07 & 25.53 & 35.80 & 58.63 & 52.25 & \textbf{62.47} & 48.99 \\
        ~~++ FLEX-Seg (Ours) & \textbf{48.51} & \textbf{28.16} & \textbf{38.34} & \textbf{59.49} & \textbf{52.48} & 61.71 & \textbf{50.07} \\
        \bottomrule
    \end{tabularx}
    \caption{Domain generalization performance with GTA as source domain (mIoU $\uparrow$ in \%). We compare our FLEX-Seg against state-of-the-art methods across both challenging conditions (ACDC, DZ) and standard datasets (CS, BDD, MV). For fair comparison, results are reported using both ResNet-101 and MiT-B5 backbones. The symbols + and ++ indicate incremental improvements, where ++ denotes methods built upon +. $^*$ means the method using ResNet-50 as the backbone, which is initialized with CLIP pretrained weights. We emphasize \textbf{best} and \underline{second best} results.}
    \label{tab:dg_results}
\end{table*}

The sampling strategy employs an adaptive threshold mechanism with sigmoid decay:

\begin{equation}
\text{threshold}(t) = \frac{1}{1 + e^{k(t - m)}}
\end{equation}
where $k$ controls the steepness of decay and $m$ is the transition midpoint. At each training iteration, we generate a random value $r \in [0,1]$. If $r > \text{threshold}(t)$, we perform loss-based sampling; otherwise, we sample randomly from the training set.

When loss-based sampling is selected, images are sampled with probability proportional to their hardness scores:

\begin{equation}
p_i = \frac{\exp\bigl(h_i(t) / \tau\bigr)}{\sum_{j} \exp\bigl(h_j(t) / \tau\bigr)}
\end{equation}
where $\tau$ is a temperature parameter controlling the sampling distribution sharpness. Higher hardness scores result in higher sampling probabilities, directing attention toward challenging cases such as adverse weather conditions or complex boundary structures.

This design ensures training stability in early stages when loss estimates may be unreliable, while progressively focusing on the most informative examples as the model matures. The result is enhanced domain generalization without requiring additional training data or model complexity.

\section{Experiments}

\begin{figure*}[t]
    \centering
    \includegraphics[width=0.95\textwidth]{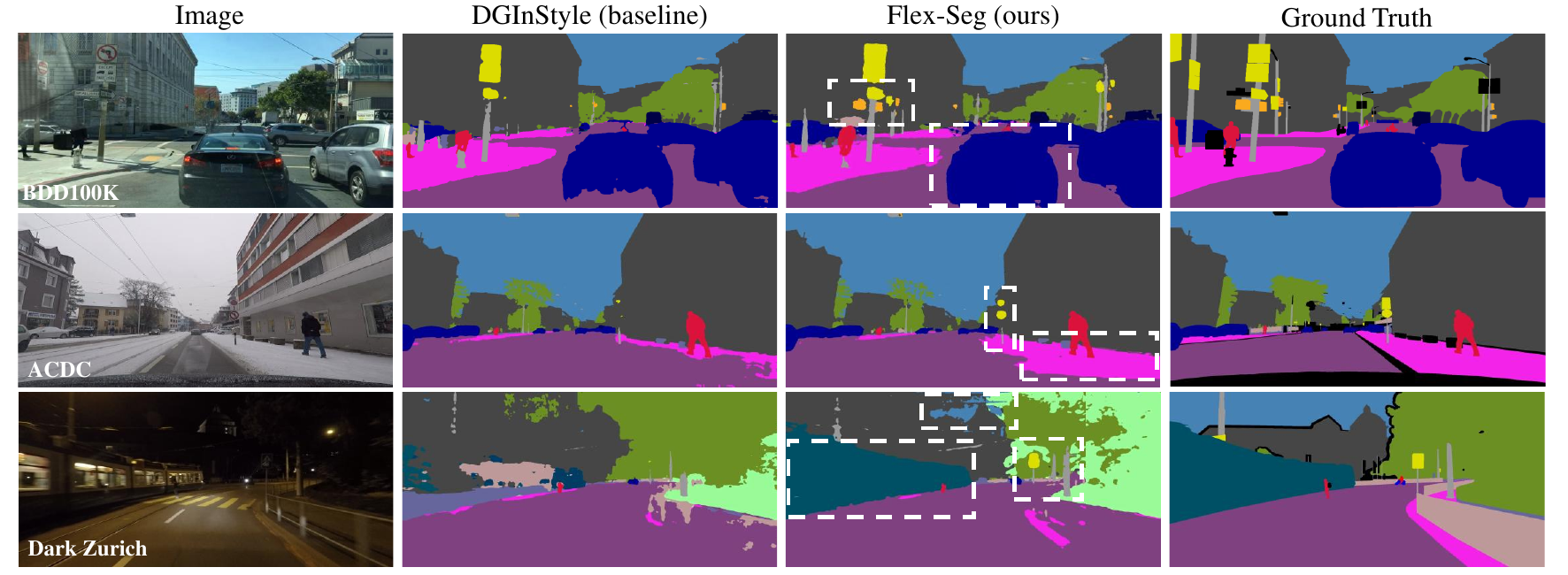}
    \caption{Qualitative comparison of segmentation results on target domains. From left to right: input images, predictions by HRDA trained with DGInStyle~\cite{jia2024dginstyle}, predictions by HRDA trained with our FLEX-Seg, and ground truth.}
    \label{fig:vis}
\end{figure*}

\subsection{Experimental Settings and Implementation Details}
\textbf{Datasets.} Following standard practice in semantic segmentation domain generalization~\cite{hoyer2023domain, hoyer2022daformer}, we use \textbf{GTA}~\cite{richter2016playing} as our source dataset, containing 24{,}966 synthetic images with 19-class annotations. For evaluation, we employ five real-world driving datasets spanning diverse conditions.
We assess performance on \textbf{Cityscapes (CS)}~\cite{cordts2016cityscapes} (500 validation images from German cities), \textbf{BDD100K (BDD)}~\cite{yu2020bdd100k} (1{,}000 images from various US locations), and \textbf{Mapillary Vistas (MV)}~\cite{neuhold2017mapillary} (2{,}000 images from worldwide locations). For challenging conditions, we include \textbf{ACDC}~\cite{sakaridis2021acdc} (406 images under fog, night, rain, and snow) and \textbf{Dark Zurich (DZ)}~\cite{sakaridis2019guided} (50 nighttime images).
This diverse collection of datasets enables thorough evaluation of our model's generalization capabilities across both standard and adverse driving conditions.

\noindent\textbf{Implementation Details.} For image generation, following~\cite{jia2024dginstyle}, we implement Rare Class Sampling (RCS) with a probability threshold of $T = 0.01$ to address class imbalance issues in the generated dataset. Specifically, semantic masks from the GTA dataset are selected using RCS, cropped to 512×512 patches centered around rare-class regions, and used to generate 10,000 diverse images covering both standard and challenging weather conditions (fog, rain, snow, and nighttime). For training, we combine these generated images with an additional 6,000 images selected from the original GTA dataset based on rare-class criteria, further ensuring balanced class representation. Our FLEX-Seg framework maintains a prototype bank of size $C \times 3 \times 256$ for GAP, where prototypes are updated using momentum factor $m=0.99$.
Our HAS strategy maintains a hardness score for each training image, updated every 50 iterations using an exponential moving average (EMA) with decay factor $\beta=0.9$. All experiments were conducted on single GPUs: RTX 3090 for DAFormer and A6000 Ada for HRDA. Other training settings (e.g., 3 random runs) follow the configurations in~\cite{hoyer2023domain}.

\noindent\textbf{Hyperparameters.} Our experiments determined the following optimal settings: GAP contrastive temperature $\tau=0.07$, UBE entropy amplification factor $\alpha=3.0$, HAS sigmoid decay parameters $k=0.05$ with sampling temperature $\tau_{HAS}=1.0$, and GAP loss weight $\lambda_{gap}=0.5$. Detailed hyperparameter sensitivity analysis can be found in supplementary material.

\subsection{Comparison with State-of-the-Art}
Table~\ref{tab:dg_results} presents comprehensive comparisons with state-of-the-art domain generalization methods. Our FLEX-Seg demonstrates consistent improvements across all evaluation settings, achieving new state-of-the-art performance.

\noindent\textbf{Challenging Domains.} On domains with adverse conditions (ACDC, DZ), FLEX-Seg shows significant gains. For ACDC, we achieve 46.56\% mIoU with DAFormer (+2.52\% over DGInStyle) and 48.51\% with HRDA (+2.44\% improvement). Notably, our method surpasses recent strong baselines FAMix (32.74\%) and SCSD (35.66\%), despite these methods using CLIP-pretrained ResNet-50. On Dark Zurich nighttime scenarios, our method reaches 29.51\% and 28.16\% mIoU for DAFormer and HRDA respectively, representing substantial improvements of +3.93\% and +2.63\%. These gains demonstrate that our approach enables accurate object delineation even under poor visibility, where precise boundary understanding becomes critical for distinguishing objects from their surroundings.

\noindent\textbf{Standard Domains.} The improvements extend to standard conditions as well. On Cityscapes, BDD100K, and Mapillary Vistas, FLEX-Seg consistently outperforms previous methods across both ResNet-101 and MiT-B5 architectures. These results demonstrate that our boundary-focused approach not only excels in adverse conditions but also enhances segmentation quality under normal driving scenarios, validating the general applicability of addressing boundary precision and uncertainty in synthetic training data.

\subsection{Qualitative Analysis}
Fig.~\ref{fig:vis} presents visual comparisons between baseline and our FLEX-Seg approach. The qualitative results reveal several key improvements: (1) More accurate boundary delineation, particularly visible in object contours such as vehicles and pedestrians; (2) Better handling of thin structures like poles and traffic signs, which are often misclassified by baseline methods; (3) Robust performance under adverse conditions including snow and nighttime scenarios, where visibility is severely limited.
These visual improvements align with our quantitative results, confirming that FLEX-Seg's focus on multi-granular boundary learning and uncertainty-adaptive weighting effectively addresses the limitations of previous approaches. The enhanced precision is especially evident in challenging weather conditions, where our method successfully delineates objects despite snow occlusion or low-light environments, while baseline methods produce fragmented or inconsistent predictions.

\subsection{Ablation Studies}
\begin{table}[t]
    \centering
    \footnotesize 
    \setlength{\tabcolsep}{2mm} 
    \begin{tabular}{ccc|ccc}
        \toprule
        \textbf{GAP} & \textbf{UBE} & \textbf{HAS} & \textbf{Avg2} & \textbf{Avg3} & \textbf{Avg5} \\
        \midrule
        - & - & - & 34.81 & 54.25 & 46.47 \\
        \checkmark & - & - & 36.33 & 55.26 & 47.69 \\
        - & \checkmark & - & 35.05 & 55.33 & 47.21 \\
        \checkmark & \checkmark & - & 36.15 & \textbf{56.07} & 48.10 \\
        \checkmark & \checkmark & \checkmark & \textbf{38.04} & 55.61 & \textbf{48.58} \\
        \bottomrule
    \end{tabular}
    \caption{Ablation studies on key components across challenging (Avg2), standard (Avg3), and all domains (Avg5).}
    \label{tab:ablation}
\end{table}

Table~\ref{tab:ablation} presents ablation studies validating the contribution of each component in FLEX-Seg. Starting from the DGInStyle baseline (34.81\% Avg2), we observe that GAP alone provides substantial improvement (+1.52\%), while UBE alone offers modest gains (+0.24\%). Combining GAP and UBE yields further improvements, reaching 36.15\% on challenging domains. The full framework with all three components achieves the best overall performance (38.04\% Avg2, 48.58\% Avg5), representing +3.23\% improvement over baseline on challenging domains.

Interestingly, while GAP+UBE achieves the highest standard domain score (56.07\% Avg3), adding HAS slightly reduces this (-0.46\%) but significantly boosts challenging domain performance (+1.89\%). This trade-off reflects HAS's adaptive sampling strategy, which prioritizes adverse examples with higher loss while maintaining balance to prevent standard domain degradation, ultimately enhancing overall robustness across diverse scenarios rather than single-environment optimization.

\section{Conclusion}

In this paper, we introduced \textbf{FLEX-Seg}, a comprehensive framework that transforms the inherent misalignment in synthetic data into an opportunity for robust domain generalization. Our approach comprises three synergistic components: Granular Adaptive Prototypes (GAP) for learning domain-invariant boundary representations across multiple scales, Uncertainty Boundary Emphasis (UBE) for dynamically emphasizing challenging regions based on prediction entropy, and Hardness-Aware Sampling (HAS) for efficient training on difficult examples.
Extensive experiments across five diverse real-world datasets demonstrate that FLEX-Seg consistently outperforms existing state-of-the-art methods. Substantial gains on challenging domains (up to +3.93\% on Dark Zurich) validate that leveraging boundary misalignment through adaptive strategies leads to superior generalization.
Our work shows that precise boundary handling of imperfect synthetic data yields better generalization than striving for perfect alignment. Future directions include exploring adaptive prototype mechanisms and extending our framework to other dense prediction tasks where boundary precision is critical.

\section{Acknowledgments}
This work was supported by the National Research Foundation of Korea (NRF) grant funded by the Korea government(MSIT)(RS-2024-00456589) and Institute of Information \& communications Technology Planning \& Evaluation (IITP) grant funded by the Korea government(MSIT) (No. RS-2025-02263277).

\bibliography{aaai2026}
\newpage
\setcounter{secnumdepth}{2}  

\setcounter{section}{0}
\setcounter{equation}{0}
\setcounter{figure}{0}
\setcounter{table}{0}

\renewcommand{\theequation}{S\arabic{equation}}
\renewcommand{\thefigure}{S\arabic{figure}}
\renewcommand{\thetable}{S\arabic{table}}

\section{Overview}

This supplementary material provides additional experimental results and analyses to support our main paper:
\begin{itemize}
    \item Section 2: Extended ablation studies
    \item Section 3: Additional motivation experiments
    \item Section 4: Additional qualitative results
\end{itemize}
\section{Extended Ablation Studies}

\subsection{Effect of UBE Amplification Factor}
\begin{figure}[H]
\centering
\includegraphics[width=\linewidth]{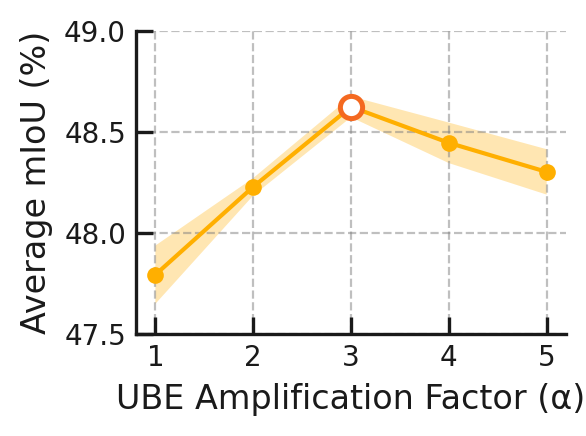}
\caption{Effect of UBE amplification factor $\alpha$ on average mIoU performance across all datasets.}
\label{fig:alpha_effect}
\end{figure}

We conduct extensive experiments to determine the optimal amplification factor $\alpha$ for our Uncertainty Boundary Emphasis (UBE) mechanism. Fig~\ref{fig:alpha_effect} shows the effect of different $\alpha$ values on the performance across all target domains.

As shown in Fig~\ref{fig:alpha_effect}, the average mIoU across all datasets increases as $\alpha$ increases, reaching a peak of 48.58\% at $\alpha = 3.0$, before gradually declining with larger values. This indicates that while dynamically emphasizing uncertain boundary regions is beneficial, excessive amplification can be detrimental. The optimal value $\alpha = 3.0$ provides the best balance between boundary uncertainty handling and overall segmentation performance.

\subsection{Effect of GAP Loss Weight}
\begin{figure}[H]
\centering
\includegraphics[width=\linewidth]{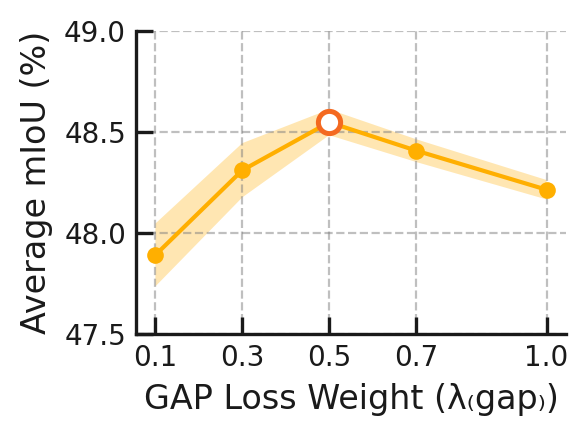}
\caption{Effect of GAP loss weight $\lambda_{gap}$ on average mIoU performance across all datasets.}
\label{fig:gap_weight_effect}
\end{figure}

We investigate the impact of different GAP loss weights ($\lambda_{gap}$) on the overall performance, as shown in Fig~\ref{fig:gap_weight_effect}.

As illustrated in Fig~\ref{fig:gap_weight_effect}, the optimal performance is achieved at $\lambda_{gap} = 0.5$ with an average mIoU of 48.58\%. The performance increases from 47.92\% at $\lambda_{gap} = 0.1$ to the peak at 0.5, then gradually decreases at higher values. This indicates that while the GAP module's contrastive learning is crucial for domain-invariant boundary representations, its contribution should be carefully balanced with the main segmentation loss to maintain overall accuracy.

\subsection{Effect of Dataset Size}
We analyze the impact of the number of generated synthetic images on model performance. Table~\ref{tab:dataset_size} shows how performance scales with different dataset sizes.

As shown in Table~\ref{tab:dataset_size}, performance steadily improves and saturates as more HAS source samples are added. Additionally, we examined how sampling evolves during training. While night and foggy conditions are sampled more frequently in the later stages due to our scheduling strategy, we observed that all conditions are sampled at relatively similar rates (around 20\% each) throughout the overall training process without bias.
\begin{table}[H]
\centering
\small
\renewcommand{\arraystretch}{1.1}
\resizebox{0.48\textwidth}{!}{%
\begin{tabular}{c|cccccc}
\hline
$N$ & 2000 & 4000 & 6000 & 8000 & 10000 & 12000 \\
\hline
Avg2 & 35.82 & 36.41 & 36.95 & 37.53 & 38.04 & 37.88 \\
Avg5 & 46.03 & 46.87 & 47.42 & 47.98 & 48.58 & 48.73 \\
\hline
\end{tabular}}
\caption{Performance according to the number of generated synthetic images.}
\label{tab:dataset_size}
\end{table}

\subsection{Effect of Sampling Temperature}
\begin{figure}[H]
\centering
\includegraphics[width=\linewidth]{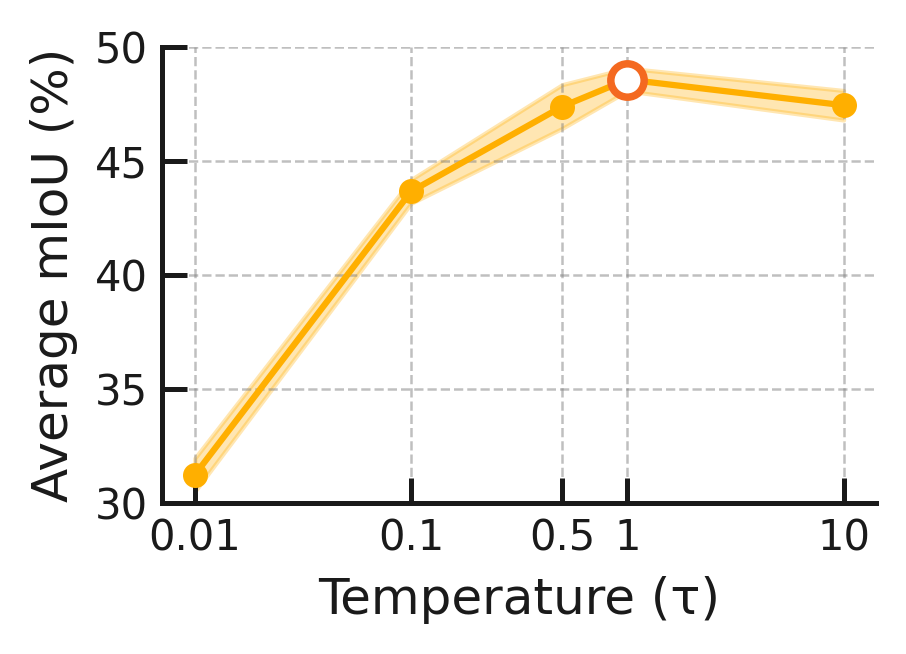}
\caption{Impact of temperature parameter $\tau$ on model performance. The optimal value of $\tau=1.0$ balances between focusing on difficult examples and maintaining training diversity.}
\label{fig:temperature_impact}
\end{figure}

The temperature parameter $\tau$ in our Hardness-Aware Sampling (HAS) strategy controls the sharpness of the sampling distribution. Fig~\ref{fig:temperature_impact} shows the effect of different temperature values on the performance.

A lower temperature ($\tau=0.01$) makes the sampling distribution extremely peaked, focusing almost exclusively on the hardest examples and neglecting other useful samples, resulting in poor performance (31.24\%). As the temperature increases to moderate values ($\tau=0.1$ to $\tau=1.0$), performance substantially improves, reaching a peak at $\tau=1.0$ (48.58\%). However, further increasing the temperature ($\tau=2.0$ to $\tau=10.0$) makes the distribution too uniform, diminishing the benefits of hardness-aware sampling and gradually reducing performance.

\subsection{Effect of Threshold Decay Strategy}
\begin{figure}[H]
\centering
\includegraphics[width=\linewidth]{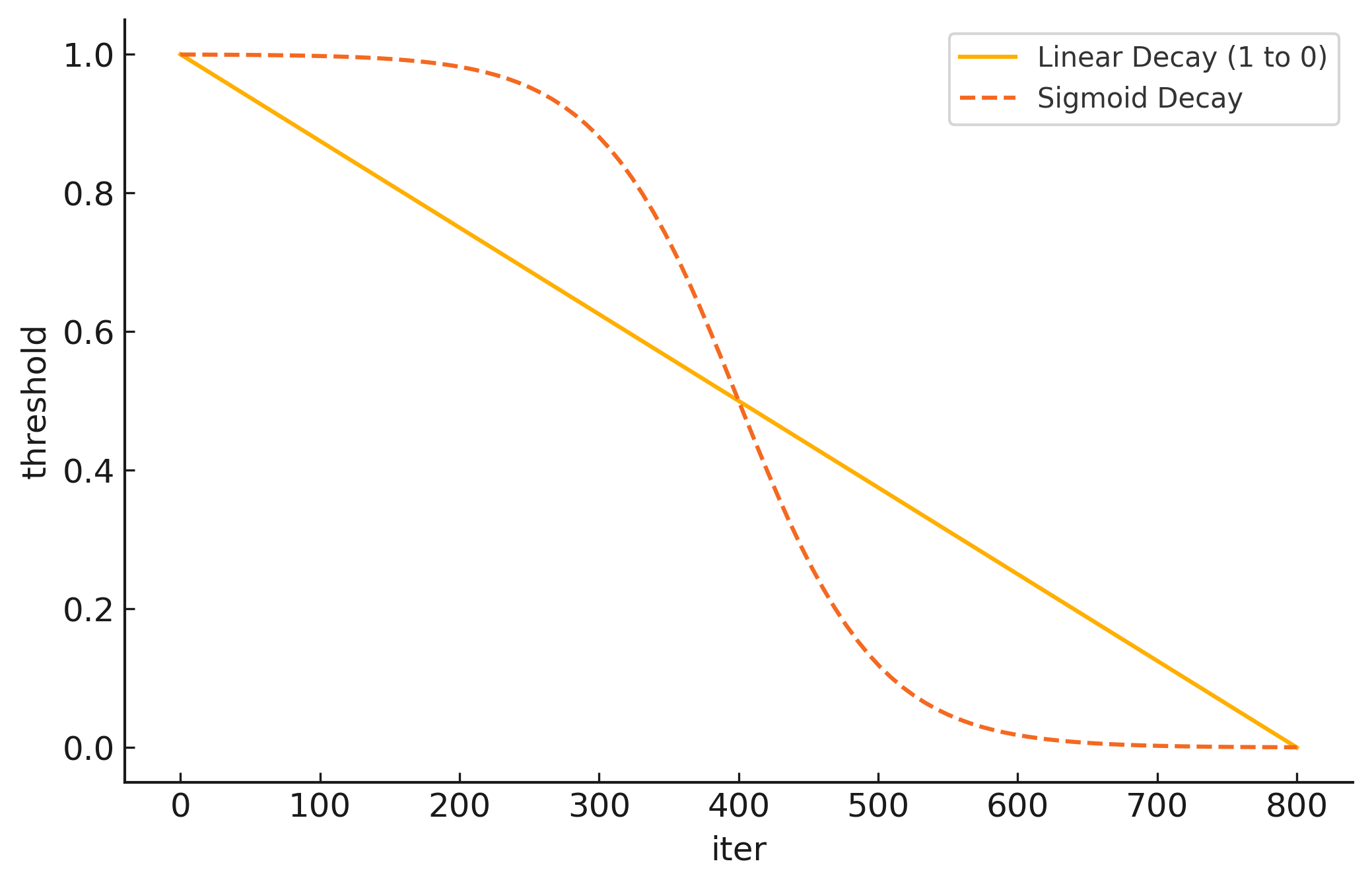}
\caption{Comparison of linear and sigmoid threshold decay strategies. Linear decay provides a more gradual transition from random to hardness-based sampling, while sigmoid decay creates a sharper transition in the middle of training.}
\label{fig:decay_strategy}
\end{figure}

We examine different threshold decay strategies for transitioning from random sampling to hardness-based sampling during training. Fig~\ref{fig:decay_strategy} compares linear decay with sigmoid decay strategies.

We explored several decay strategies including linear, sigmoid, and exponential decay. We did not adopt exponential decay as it transitions too quickly to hardness-based sampling, preventing the model from accumulating sufficient loss information during the early training stages. Table~\ref{tab:decay_strategy} compares the average performance of linear and sigmoid decay strategies across all five evaluation datasets.

Our results demonstrate that sigmoid decay strategy significantly outperforms both alternatives. Interestingly, linear decay performs worse than no decay (random sampling), suggesting that its overly gradual transition fails to effectively leverage hardness information and may even introduce instability by constantly shifting sampling priorities. The sigmoid decay provides an optimal transition pattern - maintaining random sampling in early stages to build reliable hardness estimates, then creating a decisive transition to hardness-based sampling in the middle of training when the model has accumulated sufficient loss information, and maintaining strong focus on hard examples in later stages. This adaptive transition characteristic of sigmoid decay allows it to better balance exploration and exploitation throughout the training process, leading to superior performance compared to both the static approach (no decay) and the overly gradual approach (linear decay).

\begin{table}[H]
\centering
\begin{tabular}{l|cc}
\toprule
Decay Strategy & Avg2 & Avg5 \\
\midrule
No decay (random) & 36.82 & 47.51 \\
Linear decay & 36.28 & 47.15 \\
Sigmoid decay & \textbf{38.04} & \textbf{48.58} \\
\bottomrule
\end{tabular}
\caption{Comparison of threshold decay strategies (Average mIoU \%)}
\label{tab:decay_strategy}
\end{table}

\subsection{Applicability Beyond DGInStyle}

Although our main results utilize DGInStyle for synthetic data generation, we validate the generality of FLEX-Seg by applying it to alternative diffusion-based generation methods. Specifically, we experiment with ALDM (Li et al. 2024), which employs adversarial supervision to preserve layout fidelity in layout-to-image diffusion models.

As shown in Table~\ref{tab:aldm_results}, incorporating our FLEX-Seg framework with ALDM-generated data yields consistent improvements: +1.44\% on challenging domains (Avg2) and +0.83\% across all four domains (Avg4). Since the ALDM model was pretrained on Cityscapes, we exclude it from evaluation and report Avg4 over the remaining target domains (ACDC, DZ, BDD, MV). These results demonstrate that our approach is not limited to a specific synthetic data generation method and can effectively handle boundary misalignment issues across different diffusion-based generation frameworks.

\begin{table}[H]
\centering
\begin{tabular}{lcc}
\toprule
Method & Avg2 & Avg4 \\
\midrule
ALDM baseline & 35.30 & 45.48 \\
ALDM + FLEX-Seg (Ours) & \textbf{36.74} & \textbf{46.31} \\
\bottomrule
\end{tabular}
\caption{Performance comparison using ALDM-generated synthetic data}
\label{tab:aldm_results}
\end{table}

\section{Additional Motivation Experiments}

We investigated three alternative strategies for handling noisy or ambiguous boundary regions in synthetic training data, in addition to the boundary enhancement approach presented in the main paper.

\subsection{Alternative Boundary Handling Strategies}

The boundary region $B$ is identified by applying morphological operations to the semantic masks:
\begin{equation}
B = \text{Dilate}(M, k_d) - \text{Erode}(M, k_e),
\end{equation}
where $M$ indicates the training mask, and $k_d$, $k_e$ are dilation and erosion kernel sizes respectively.

\begin{enumerate}
    \item \textbf{Ignoring Boundary Regions:} 
    \begin{equation}
    \mathcal{L}_{\text{ignore}} = \sum_{(x,y) \notin B} \mathcal{L}_{\text{CE}}(x,y),
    \end{equation}
    where $\mathcal{L}_{\text{CE}}(x,y)$ is the cross-entropy loss at pixel $(x,y)$.
    
    \item \textbf{Threshold-based Boundary Adjustment:} 
    \begin{equation}
    \begin{aligned}
    \mathcal{L}_{\text{threshold}} = &\sum_{(x,y) \in B} \min(\tau, \mathcal{L}_{\text{CE}}(x,y)) \\
    &+ a \cdot \max(0, \mathcal{L}_{\text{CE}}(x,y) - \tau) \\
    &+ \sum_{(x,y) \notin B} \mathcal{L}_{\text{CE}}(x,y),
    \end{aligned}
    \end{equation}
    where $\tau$ is a predefined threshold and $a < 1$ is a scaling factor.

    \item \textbf{Reducing the Effect of the Boundary Regions:} 
    \begin{equation}
    \mathcal{L}_{\text{reducing}} = \sum_{(x,y)} W_{\text{red}}(x,y) \cdot \mathcal{L}_{\text{CE}}(x,y),
    \end{equation}
    where $W_{\text{red}}(x,y) = \gamma < 1$ if $(x,y) \in B$, and $1$ otherwise.
\end{enumerate}

\subsection{Visualization and Analysis}

\begin{figure*}[!htbp]
\centering
\includegraphics[width=\textwidth]{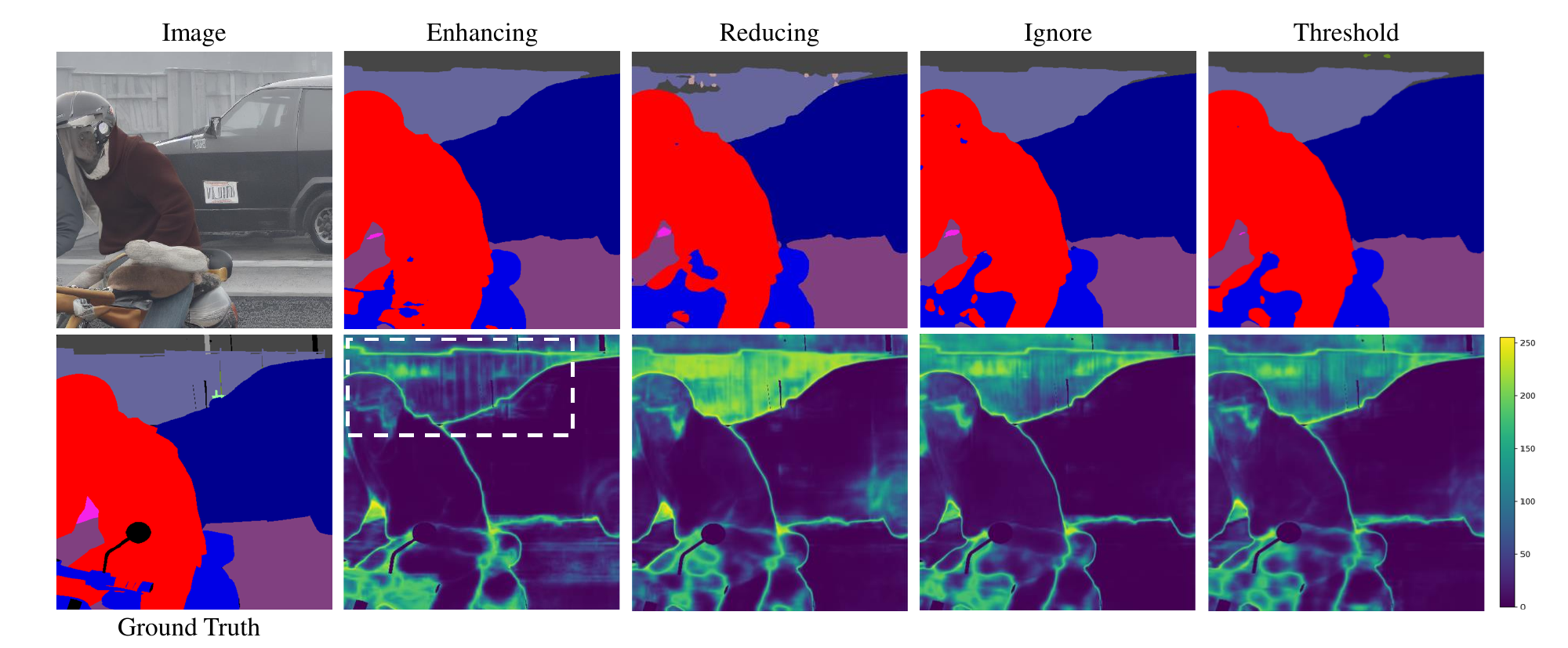}
\caption{Visualization of different boundary handling strategies on simple boundary structures. Alternative strategies produce higher entropy maps and less accurate segmentation results.}
\label{fig:obser1}
\end{figure*}

\begin{figure*}[!htbp]
\centering
\includegraphics[width=\textwidth]{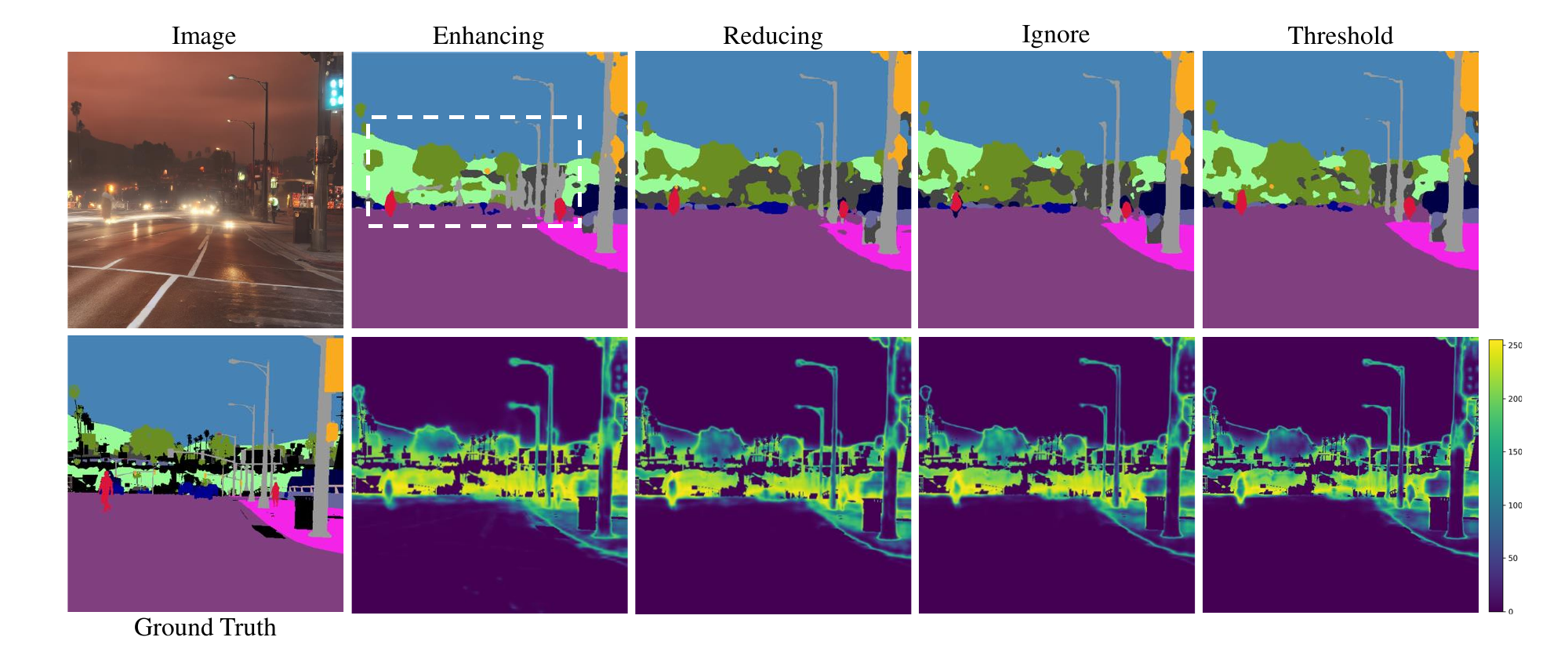}
\caption{Boundary handling strategies under poor lighting conditions. Alternative strategies struggle with complex object boundaries.}
\label{fig:obser2}
\end{figure*}

\begin{figure*}[!htbp]
\centering
\includegraphics[width=\textwidth]{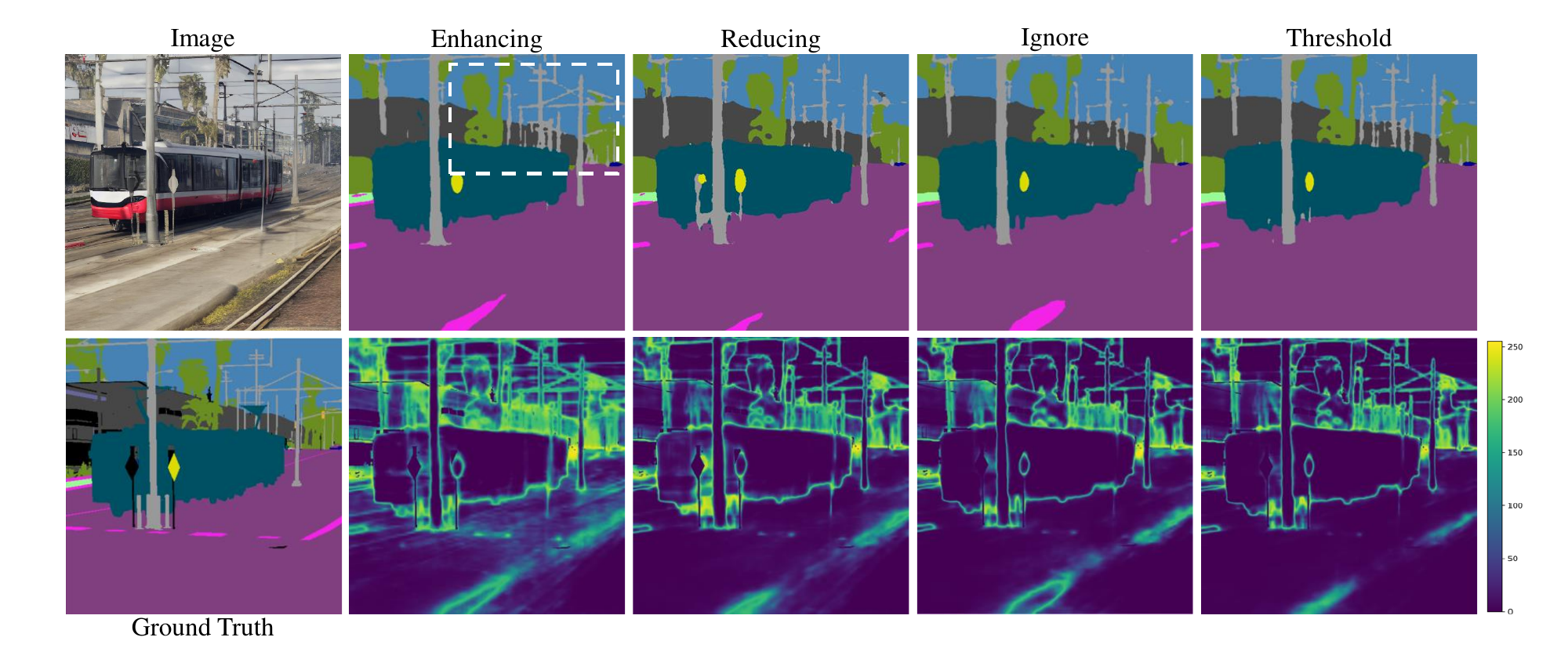}
\caption{Boundary handling strategies in structurally complex environments. Alternative strategies fail to maintain clear separation between semantic classes.}
\label{fig:obser3}
\end{figure*}

\begin{figure*}[!htbp]
\centering
\includegraphics[width=\textwidth]{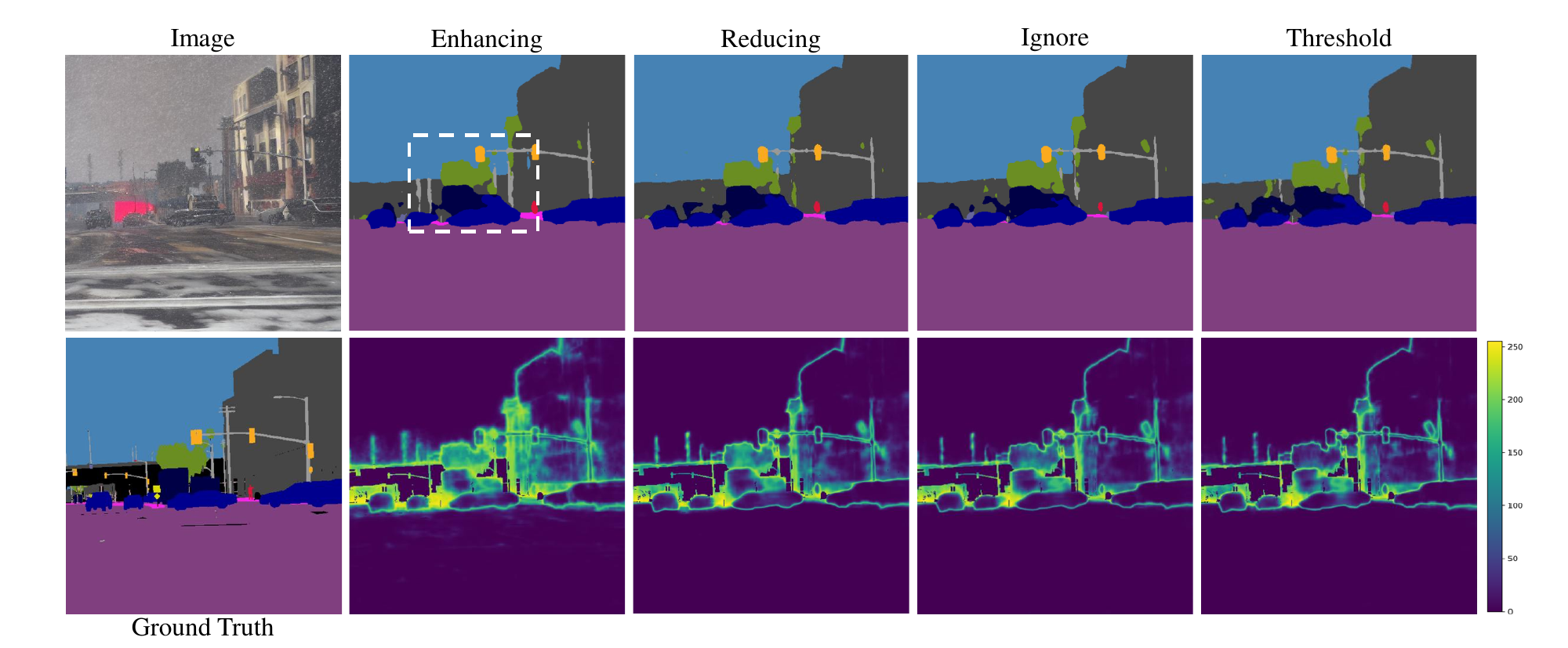}
\caption{Boundary handling strategies in structurally complex environments. While alternative strategies fail to maintain clear separation between semantic classes, the boundary enhancement approach successfully captures misaligned regions by maintaining high entropy at uncertain boundaries, enabling robust segmentation despite imperfect synthetic data alignment.}

\label{fig:obser4}
\end{figure*}

Figures~\ref{fig:obser1} through \ref{fig:obser4} visualize the effectiveness of different boundary handling strategies. Our experiments reveal that ignoring boundaries leads to incomplete object delineation, threshold-based adjustment shows inconsistent results, while reducing boundary effects causes blurred predictions at object edges. In contrast, the boundary enhancement approach (presented in the main paper) consistently produces the most accurate segmentation results by maintaining low entropy in clear interior regions while preserving appropriately high entropy at ambiguous boundary areas. This balanced uncertainty distribution indicates that the model makes confident predictions where visual cues are clear, yet remains appropriately uncertain at challenging boundaries rather than making overconfident mistakes. This superior performance validates our decision to build FLEX-Seg upon the boundary enhancement principle, which effectively guides the model to focus on challenging boundary regions while maintaining robust predictions across the entire image.

\section{Additional Qualitative Results}

\subsection{Synthetic Data Examples}
\begin{figure}[H]
\centering
\includegraphics[width=\linewidth]{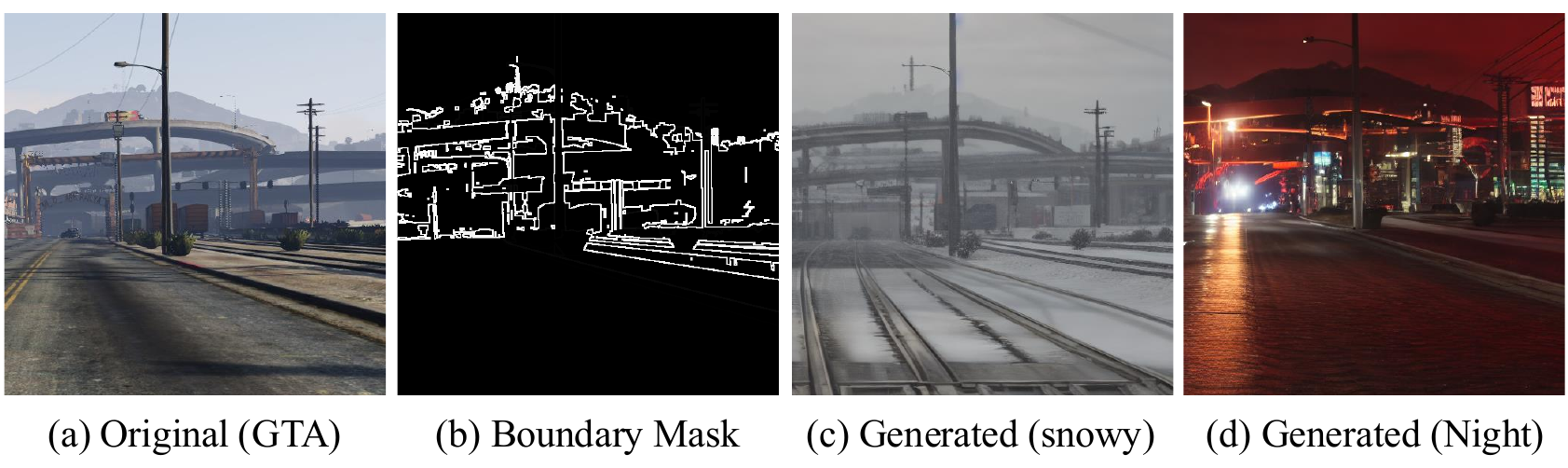}
\caption{Examples of synthetic images with corresponding semantic masks and extracted boundary regions used in our training process.}
\label{fig:synthetic_examples}
\end{figure}

Figure~\ref{fig:synthetic_examples} shows examples of synthetic images generated using diffusion models along with their corresponding semantic masks and boundary regions.

\subsection{Further example predictions}
\begin{figure*}[!htbp]
\centering
\includegraphics[width=0.85\textwidth]{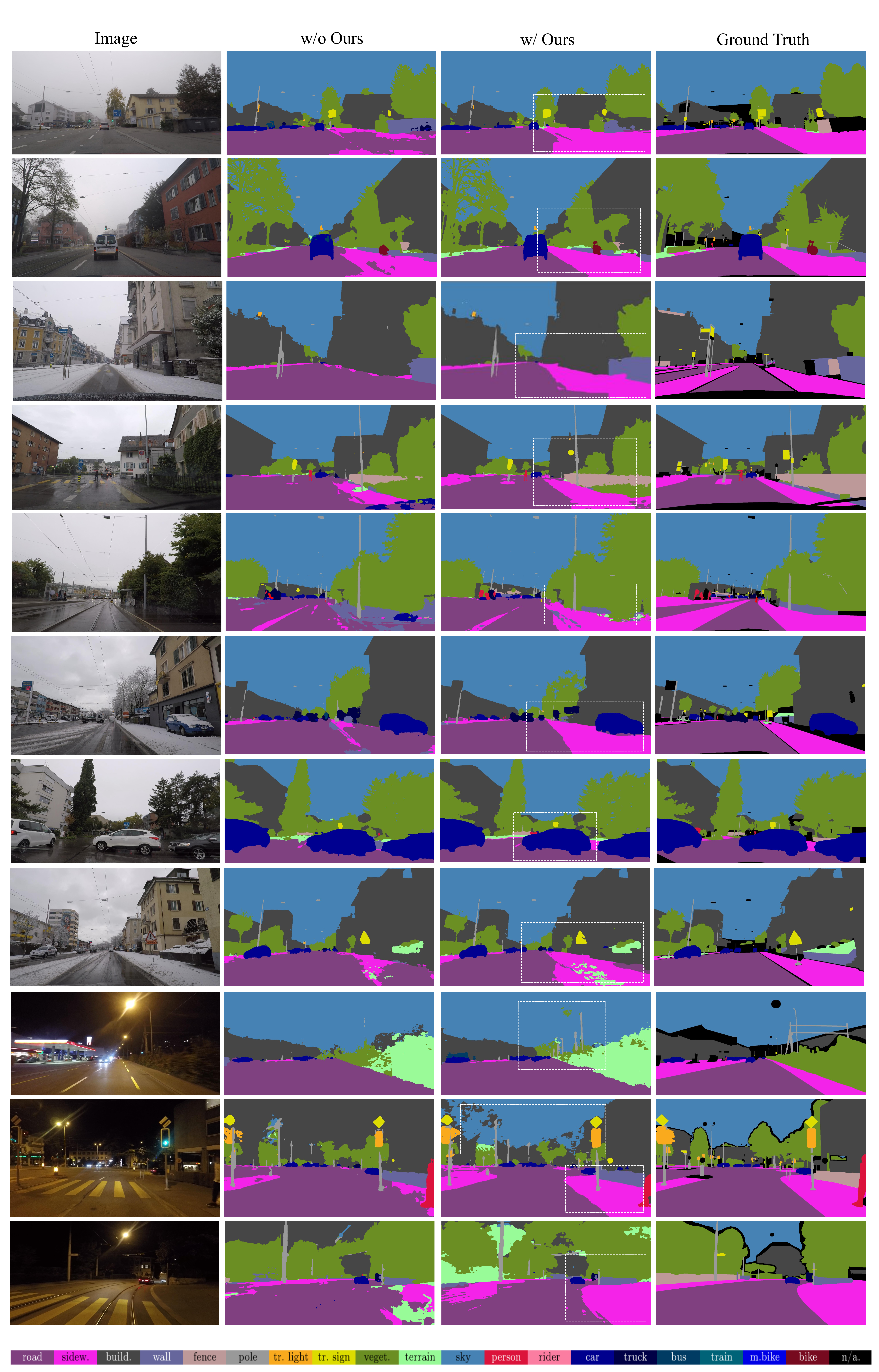}
\caption{Qualitative results on challenging datasets (ACDC, Dark Zurich). Our method maintains accurate segmentation under adverse weather conditions including fog, rain, snow, and nighttime scenarios.}
\label{fig:vis_adverse1}
\end{figure*}

\begin{figure*}[!htbp]
\centering
\includegraphics[width=0.85\textwidth]{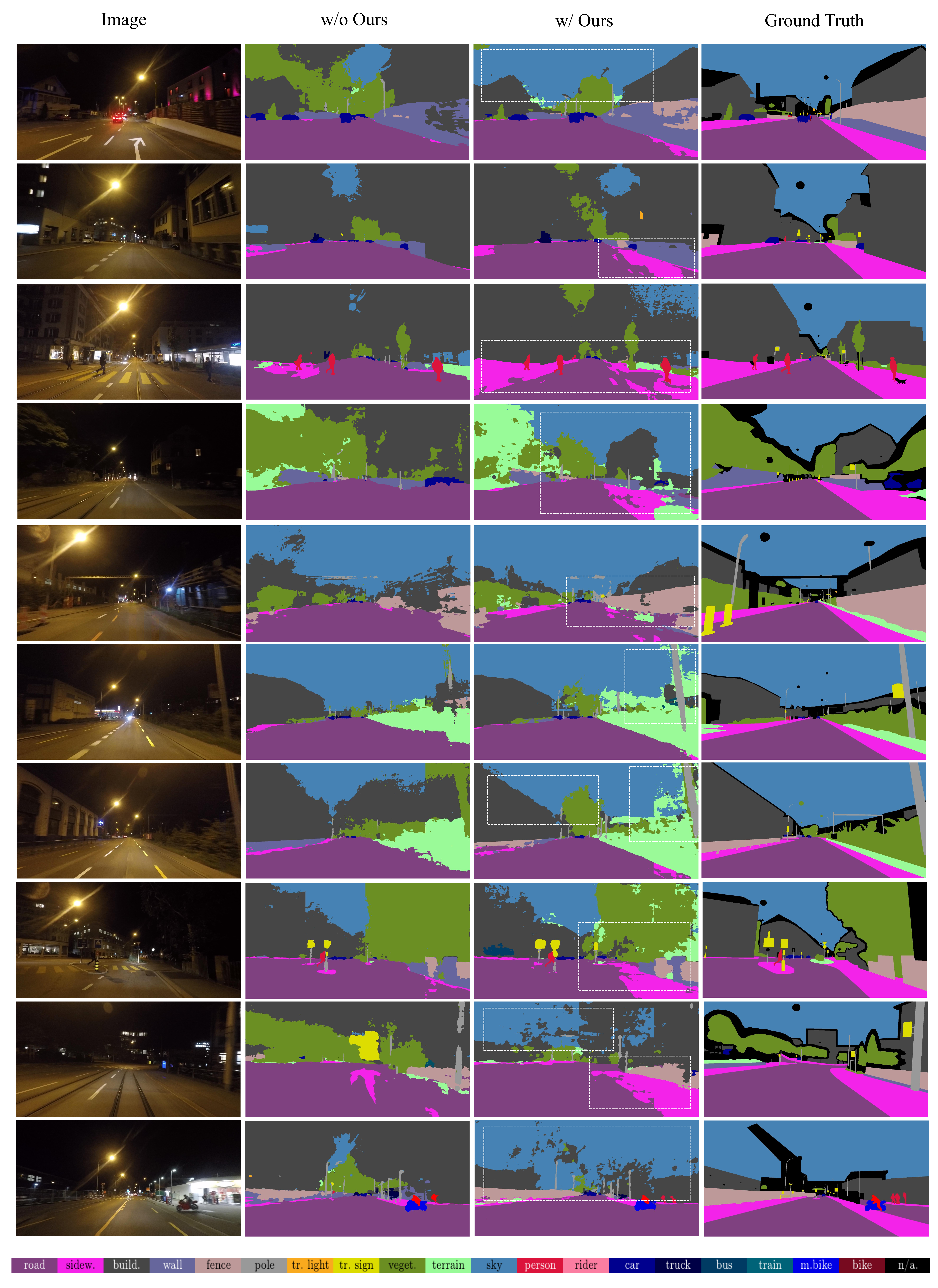}
\caption{Qualitative results on challenging datasets (ACDC, Dark Zurich). Our method maintains accurate segmentation even under challenging nighttime conditions.}
\label{fig:vis_adverse2}
\end{figure*}

\begin{figure*}[!htbp]
\centering
\includegraphics[width=0.85\textwidth]{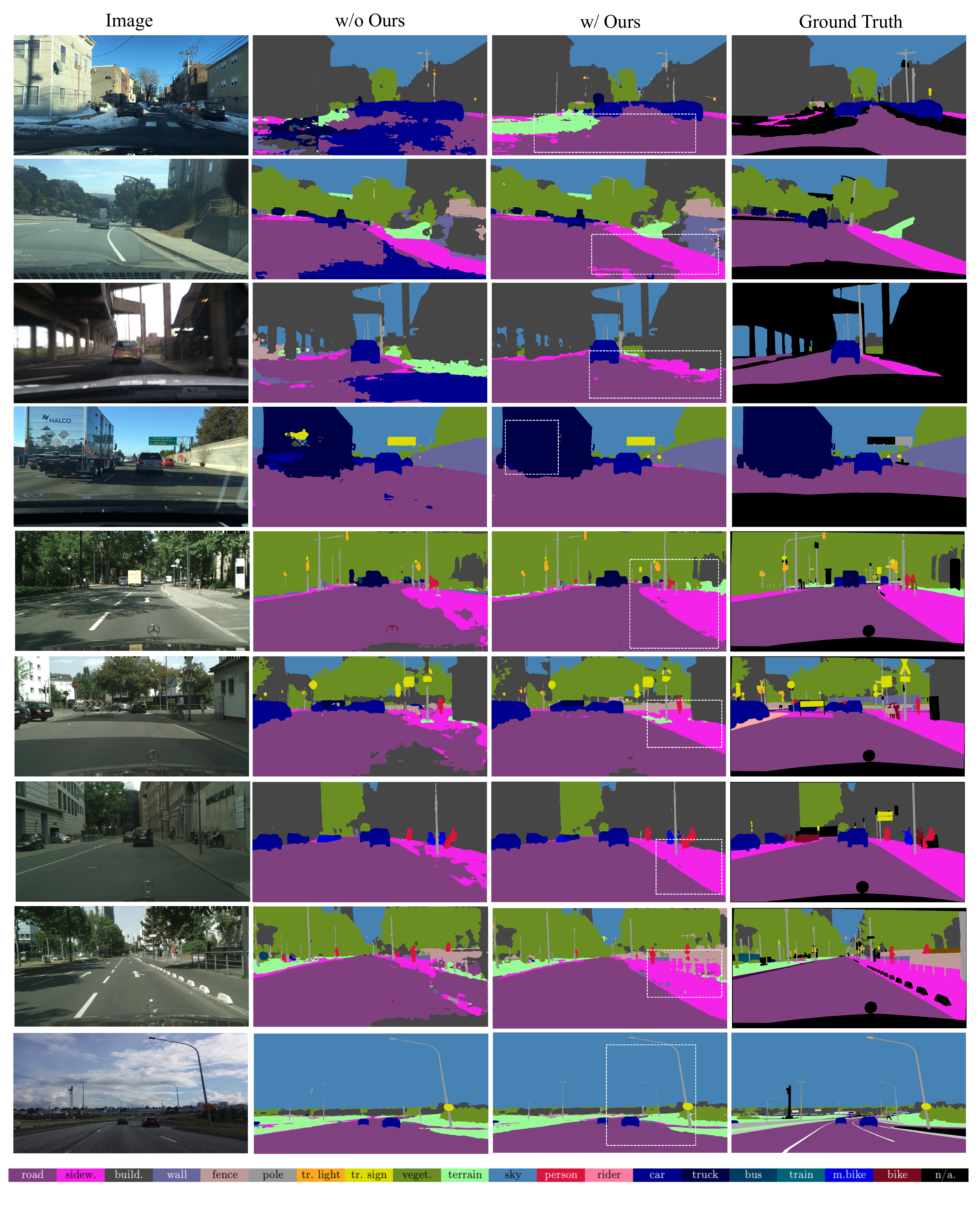}
\caption{Qualitative results on standard datasets (Cityscapes, BDD 100K, Mapillary). Our method achieves precise boundary delineation in standard urban driving scenarios.}
\label{fig:vis_normal1}
\end{figure*}

\begin{figure*}[!htbp]
\centering
\includegraphics[width=0.85\textwidth]{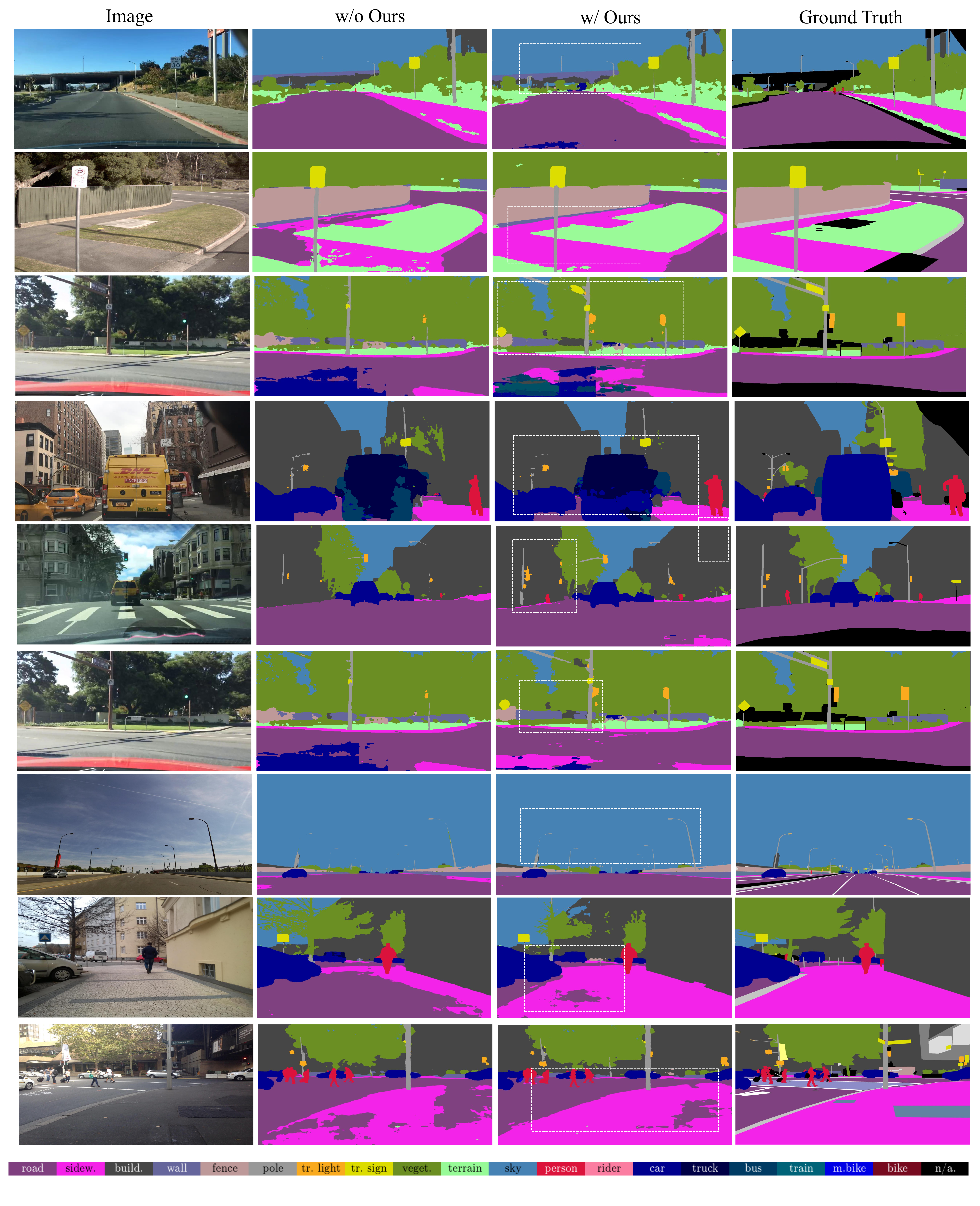}
\caption{Qualitative results on standard datasets (Cityscapes, BDD 100K, Mapillary). Our method achieves precise boundary delineation in standard urban driving scenarios.}
\label{fig:vis_normal2}
\end{figure*}

Figures~\ref{fig:vis_adverse1} and \ref{fig:vis_adverse2} show qualitative comparisons on challenging domains (ACDC and Dark Zurich), while Figures~\ref{fig:vis_normal1} and \ref{fig:vis_normal2} present results on standard domains.

\end{document}